\def\year{2022}\relax
\documentclass[letterpaper]{article} 
\usepackage{aaai22}  
\usepackage{times}  
\usepackage{helvet}  
\usepackage{courier}  
\usepackage[hyphens]{url}  
\usepackage{graphicx} 
\def\UrlFont{\rm}  
\usepackage{natbib}  
\usepackage{caption} 
\DeclareCaptionStyle{ruled}{labelfont=normalfont,labelsep=colon,strut=off} 
\usepackage{algorithm}
\usepackage{algorithmic}
\usepackage{tabulary}
\usepackage{newfloat}
\usepackage{listings}
\floatstyle{ruled}
\newfloat{listing}{tb}{lst}{}
\floatname{listing}{Listing}

\def\name{{NAERS}}
\title{Continuous Learning Based Novelty Aware Emotion Recognition System}
\author {
    Mijanur Palash,
    Bharat Bhargava
}
\affiliations {
    Department of Computer Science, Purdue University, USA\\
    \{mpalash, bbshail\}@purdue.edu
}

\date{November 2021}

\begin{document}

\maketitle

\section{Abstract}
Current works in human emotion recognition follow the traditional closed learning approach governed by rigid rules without any consideration of novelty. Classification models are trained on some collected datasets and expected to have the same data distribution in the real-world deployment. Due to the fluid and constantly changing nature of the world we live in, it is possible to have unexpected and novel sample distribution which can lead the model to fail. Hence, in this work, we propose a continuous learning based approach to deal with novelty in the automatic emotion recognition task.

\section{Introduction}\label{intro}
Understanding emotions properly is an important requirement for meaningful human communication. In addition, emotional state affects things like how a person drives a car, how a kid learns in the classroom or how police interact with a criminal on the road etc. Automatic emotion recognition has widespread usage in different areas such as human-computer interactions, law enforcement and surveillance, interactive gaming, consumer behaviour analysis, customer service, education, and health care etc. 

Researchers in psychology categorized basic human emotions into anger, happiness, sadness, disgust, fear, contempt and surprise~\cite{patel2020facial} classes. Human expresses these emotions using various verbal and non-verbal cues. Facial expression analysis is one of the major ways to identify the underlying emotion. Apart from that, various important conclusions for emotion can be drawn from posture and gait. Similarly, speech, writing, brain scan, EEG signal etc. also convey useful emotional information. 

The advancement of machine learning and deep learning enables researchers to design more accurate classifiers for automatic emotion recognition systems. Researchers nowadays frequently use various ML and DL techniques to detect and classify emotions. Popular models include variations of support vector machine (SVM), deep belief network (DBN), ensemble learning, deep neural network (DNN) and convolutional neural network (CNN) etc. These models are trained on various datasets collected from web scrapping or created using volunteers and involve somebody to manually label the data samples. More details about some of the datasets are discussed in section~4. 

However, this closed learning approach faces challenges in real-world situations where an unexpected sample may appear at any moment. A perfect machine learning model requires a perfect training dataset that truly represents all possible real-world situations. However, creating an all-inclusive dataset like this is very hard. Moreover, an emotion detection model trained on acted datasets where the margin between true and false samples are wide and clear may have a hard time correctly identifying emotion in a situation where someone is intentionally obfuscating his emotional state and hence the margin is thin and blurry. Another type of challenge arises when the model encounters a different type of emotion class absent in the training dataset as most of the datasets do not cover all of the emotion classes. In this work we define these types of situations as a novelty. 

However, there is a lack of proper focus on handling novelty in current emotion recognition research. Researchers are more interested in coming up with newer models and newer datasets to get higher accuracy on the test data. However, we also need to look for answers to the questions such as how to detect a novelty, how to characterize it and how to adapt the systems to handle the novelties etc. 

The main contributions of this work are:
\begin{enumerate}
    \item We present the novelty aspects of the automatic emotion recognition task.
    \item We present a system that addresses novelty in a continuous long learning manner for emotion recognition.
    \item We present an emotion recognition classifier that detects emotion from facial image
    \item We present a novelty detector that works in parallel with the classifier to detect novelty
    \item We present experimental results and discuss potential next steps for this work.

\end{enumerate}

\section{Background and Related Work}\label{background}

Convolutional Neural Network (CNN)~\cite{lecun} is a deep learning architecture widely used in image analysis. A CNN typically consists of convolutional, pooling and fully connected layers stacked together. The convolution can be defined as a repeated filter operation to an input that results in a feature map. For image classification, we can say CNN highlights important parts of the image to differentiate the classes. 

Jadhav et. al.~\cite{jadhav} used a convolutional neural network (CNN) to detect emotion from the facial expressions on the FER-2013 dataset. They achieved only 63\% accuracy on the test dataset. This lower accuracy can be attributed to the simplicity of their model. 

Gan et. al.~\cite{gan} achieved improved accuracy on the FER-2013 dataset using ensemble CNN and a novel label perturbation strategy. Due to the enhanced discrimination ability of the ensemble method, they received higher accuracy of 73.7\%. 

Dhankar et. al.~\cite{dhankhar} used the ensemble method and transfer learning with VGG16 and RESENT-50 to overcome the limitations of the basic CNN method. However, their work did not achieve the best result and only scored 67\% accuracy on the FER-2013. This may be attributed to a sub-optimal hyperparameter tuning.

Besides CNN, the support vector machine (SVM) is also popular in facial emotion recognition due to its lightweight architecture compared to CNN. Support Vector Machine (SVM) is a type of machine learning model which tries to identify the maximum margin plane between the classes. Datta et. al.~\cite{datta} used SVM for emotion classification. Kurup et. al.~\cite{kurup} used the deep belief network (DBN) technique for emotion classification.

\section{Datasets}\label{datasets}
In literature, several datasets have been proposed for emotion recognition. A summary of notable datasets is presented in table \ref{datasets}. 
 
\begin{table}[htb]
\centering
\caption{Different datasets for automatic emotion recognition. Here, N: Neutral, S: Sadness, Sr: Surprise, H: Happiness, F: Fear, A: Anger, B: Boredom, P: Puzzlement, Ax: Anxiety, C: Contempt and D: Disgust.}
\begin{tabular}{|l|l|l|p{0.2\linewidth}|p{0.2\linewidth}|p{0.2\linewidth}|}
\hline
\bf{Name}& \bf{\# Items}   & \bf{Type}& \bf{Setting}  &\bf {Classes} \\ 
 \hline
 CK+ & 593&Video&Posed \& spontaneous&N, S, Sr, H, F, A and D\\
 \hline
   FER-2013& 32,298 &   Image& Posed& N, S, Sr, H, F, A and D \\
   \hline
  Emotic & 23,571  &   Image& Wild  &N, S, Sr, H, F, A, D and 19 other classes \\
  \hline
   AffectNets& 450,000  &    Image& Wild  & N, S, Sr, H, F, A, D and C \\
   \hline
   CAER-S & 70,000     & Image& TV shows&N, S, Sr, H, F, A and D\\
   \hline
      FABO & 206     & Video& Posed&N, S, Sr, H, F, A, B, P, Ax and D\\

\hline
\end{tabular}
\label{datasets}
\end{table}

\section{Proposed System: \name} 
\begin{figure}[htb]
\centering
   \includegraphics[width=.9\linewidth]{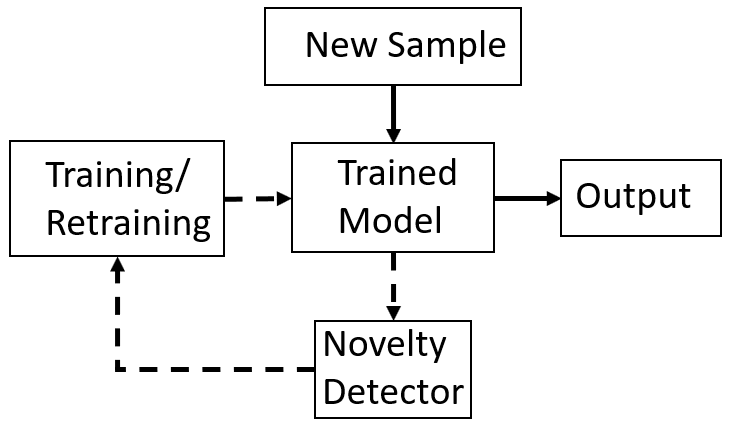}
   \caption{Proposed NAERS system architecture}
   \label{fig:noveltySys}
\end{figure}

In this section, we discuss our proposed\space\name\space which stands for \textbf{N}ovelty \textbf{A}ware \textbf{E}motion \textbf{R}ecognition \textbf{S}ystem. Figure~\ref{fig:noveltySys} shows a high-level diagram of different parts of this system.  

\subsection{Input} The input of this system is the samples of RGB frames with human subjects. This can be a static photo or a video frame. Live video can be also used for continuous recognition. In that case, we sample video frames in a suitable frequency such as 10 FPS (frames per second).

\begin{figure}[htb]
\centering
   \includegraphics[width=0.9\linewidth]{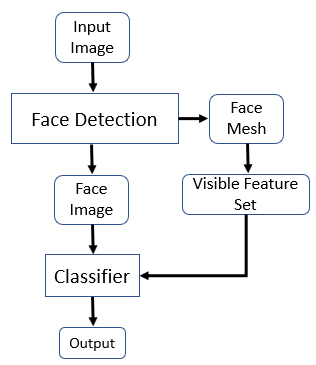}
   \caption{Proposed emotion recognition model architecture}
   \label{fig:sysD}
\end{figure}

\subsection{Emotion Recognition (ER) Model }\label{er}
\subsubsection{Face and Background Separation} We use the facial image as our main source of the emotional information. Therefore, We separate the face from the background of the input frame. while the face is used for facial emotion recognition, the rest of the body and background are used for novelty detection (section~\ref{novelty_detection}). For facial area detection, we use the Blazeface face and body detection tool~\cite{mediapipe}. It can detect multiple landmarks on the face. 

\subsubsection{Visible Features Generation}~\label{visf}
We create a face mesh using~ Blazeface face mesh generation tool~\cite{mediapipe}. From this mesh, we generate the features shown in the table~\ref{vis_feat}. While shouting or laughing, people generally open their mouths resulting in a larger separation between the lips. Similarly, a surprised person will have big open eyes resulting in a bigger area. So these features convey meaningful information valuable to us. We also take the angle values between different face parts such as eyes and nose to capture facial distortion. They are normalized with face height to compensate for body size variations. 

\begin{table}[]
\centering
\caption{Visible features}
\begin{tabular}{|l|l|}
\hline
\bf{Feature Type}  &\bf{ Feature Description}  \\ 
 \hline

  Width& Left eye\\
   
   &Right eye\\

& Mouth\\
\hline
Height&Right eye\\

&Left eye\\

& Right eye\\

&Mouth\\
\hline
 Distance & Left and right eyes\\
   \
   &Eyes to brows\\

& Eyes to mouth\\

& Eyes and nose\\

&Nose and mouth\\
\hline
Angle &Left eye with right eye and mouth\\

&Right Eye with left eye and mouth\\

&Mouth with both eyes\\

&Mouth with both eyes\\
\hline

\end{tabular}
\label{vis_feat}
\end{table}

\subsubsection{Deep Features Generation}~\label{deepf}
\begin{figure}[htb]
\centering
   \includegraphics[width=.9\linewidth]{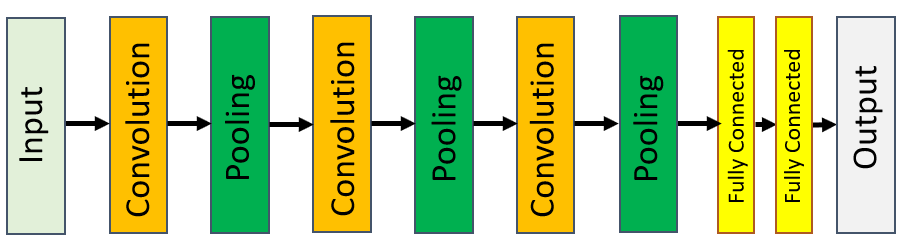}
   \caption{Proposed Convolutional Neural Network (CNN) architecture for deep features generation}
   \label{fig:cnn_model}
\end{figure}

The emotion recognition system of \name \space uses the convolutional neural network to generate deep features from the input images. These deep features are then concatenated with the visible features (\ref{visf}) to generate the final feature vector for the classification model. For deep features generation, we experiment with three types of CNN: 
\begin{itemize}
    \item Regular CNN: We use two convolutional and two fully connected layers as shown in figure~\ref{fig:cnn_model}. Input images are scaled to $226\times226$. We use a filter size of $2\times2$ and a stride of 1. The downsampling layers used $2\times2$ max-pooling. 
    \item Transfer Learning: We use the transfer learning approach in this type using Resnet-50 (~\cite{he2015deep}) trained with ImageNet. Resnet-50 is a 50 layers deep neural network.  
    \item Ensemble CNN: In ensemble learning several primary models (weak learners) are trained individually on the data set.  We use multiple regular CNN models as our weak learners. The model variations come from the variation of the initial weights. We used the average of the outputs from the model as the final output
\end{itemize}

We do not use the classification output provided by these deep feature generation models. Instead, we use the representation of the input samples at the output of the last layer before the linear layers as deep features. 

\subsubsection{Emotion Classifier} Both the visible and deep feature sets are concatenated to form a final feature vector. We pass this feature vector through a 3-layer deep neural network (DNN) which provides class probabilities for the input sample. In the deep layers, we use the Relu activation function and the final layer output is passed through a softmax function to generate the probability values. The model is trained in the mini-batch gradient descent fashion. Different mini-batch sizes, selection of an optimizer from options such as Momentum, Adam and Nesterov, and learning rate are the hyperparameters of the model.  

\subsection{Novelty Detector}\label{novelty_detection}
We propose a new algorithm for the novelty detector to detect novel and unexpected data samples. It relies on available multimodality of the data samples:
\subsubsection{Additional modality} As we discussed in section~\ref{intro}, human emotion can be identified from not only their face but also from posture and gesture extracted from the same input sample. For example, a confident person would appear relaxed while an afraid person would appear tensed and shrunk. For the input sample, we also analyze posture. If both facial and posture network output does not concur we tag that sample as a potential novelty. For posture based detection, we remove the facial area from the input. We use a kinematic human body model to represent the body posture and identify key body joints on the frame using the Blazepose tool. By following a similar idea we used in section~\ref{er}, we create a visible feature set from different joint angles and distances. We use three CNN models (section~\ref{deepf}) to generate deep features. The final feature set is created by concatenating visible and deep feature sets. The classifier is also a similar 3 layer deep neural network. For the brevity of the manuscript, we do not provide the details of the posture network here.  
\subsubsection{Context} Context is important for novelty detection. A tiger in a forest is a normal sample but the same tiger in a busy city street indicates something is wrong. We create context from the background of the image. At first human body and face is removed from the input frame. Then we compare sample background with the training background distribution using clustering techniques. If a significant difference in the background is detected we tag the input sample as a novelty.

\subsection{Retraining} Retraining is done periodically and initiated when a sufficient number of novelty samples are encountered. During the retraining phase, we explore the clustering options to check whether a new class label is warranted or not. if a large number of novelty samples gather around together that indicates the requirement of a new class label. The most probable reason for that may be an important class label being missing from the training data. So we add this new class label. Right now re-labelling is done by human operators. We are exploring other options possibly without human involvement as future work. For the situation where novelty samples do not indicate new class label requirement, rather a mismatch between face and posture modalities, we update their class label with more precise manual labelling and increase their weight in the training sample distribution in Adaboost fashion.

\section{Experimental Results}\label{result}
In this section, we discuss some of the experiments we have performed so far. 

\subsection{Performance of Emotion Recognition Model} We got 68\%, 71.5\% and 75.8\% test accuracies for the three versions of our model with different CNN models on  FER-2013 dataset. In addition, we compared our results with similar works on this dataset as shown in table~\ref{fer1}. Our approach with the ensemble of CNN outperforms the other two models and other existing works. We attribute this improvement to the variance reduction achieved by the ensemble operation. For the rest of this manuscript, we only consider ensemble CNN while talking about our emotion recognition model.

\begin{table}
\begin{tabular}{|p{0.38\linewidth}|p{0.22\linewidth}|p{0.22\linewidth}|}
\hline
 \bf{Author}& \bf{Model}   &\bf{ Accuracy(\%)}  \\ 
 \hline
   \cite{mollahosseini}&  CNN & 66.0  \\
   \hline
   \cite{dhankhar} & RESNET-50    &67.2 \\
   \hline
   \cite{renda} &  Ensemble& 71\\
   \hline
   \cite{gan}&  Ensemble& 73.73\\
   \hline

   This work&Ensemble CNN &  75.8\\
\hline

\end{tabular}
\caption{Performance comparison of different emotion recognition systems on FER-2013 dataset }
\label{fer1}
\end{table}

\begin{figure}[htb]
\centering
   \includegraphics[width=.9\linewidth]{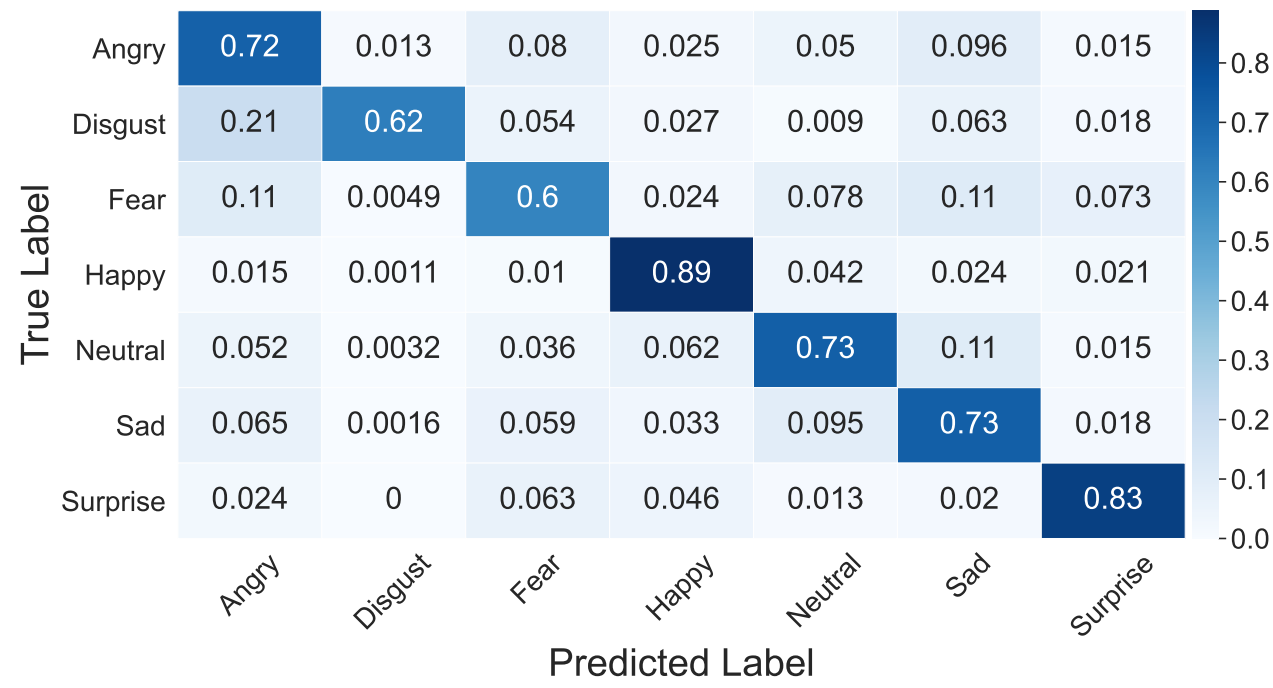}
   \captionof{figure}{Confusion matrix of our emotion recognition model for FER-2013 dataset.}
   \label{conf_mat}
\end{figure}

Figure~\ref{conf_mat} shows the confusion matrix for our emotion recognition model for the FER-2013 dataset. Here actual emotion labels are plotted along the vertical while predicted labels are along the horizontal axis. An entry `$C_{ij}$' of row `$i$' and column `$j$' represents the fraction of total samples having the true label of row `$i$', and predicted label of column `$j$'. The diagonal values are the accuracy of the class and a higher value is represented by a darker colour. We notice the happiness class has higher accuracy implying it is easier to detect. For fear, the accuracy is lower and that's because the system confuses fear with sadness. These two are closely related emotions that may appear together and people may actively try to conceal their fear with other similar emotion classes. 

The results of the facial emotion recognition model on some other datasets are shown in table~\ref{fer2}. Our model does well while comparing other works in FER-2013, AffectNet and CAER-S datasets. For the FABO dataset, we outperformed the accuracy of 76.4\% reported by the authors of the dataset~\cite{gunes2007bi} by a wide margin. For the CK+ dataset, we found the best-reported result to be 98.57\% accuracy ~\cite{kurup}. Our accuracy of 98.5\% is close to it.

\begin{table}
\begin{tabular}{|c|c|}
\hline

\centering
 \bf{Dataset} &\bf{ Accuracy(\%)}    \\ 
 \hline
  CK+& 97.5 \\
  \hline
   FER-2013& 75.8 \\
   \hline
   AffectNets&66.5 \\
   \hline
   CAER-S & 72.4\\
   \hline
   FABO & 96.1    \\
\hline

\end{tabular}
\caption{Performance of our emotion recognition model on different related datasets}
\label{fer2}
\end{table}

\subsection{Novelty Detection Experiment}
As mentioned earlier in section~\ref{novelty_detection}, for novelty detection we used posture as an additional modality and check the mismatch between facial based and posture based recognition. For training the posture emotion recognition model, We used several posture emotion recognition datasets such as FABO, Emotic and CAER-S.  

In the FABO dataset, our posture model achieved 83.1\% accuracy.  
Results from several other recent posture based work on FABO dataset are shown in table~\ref{p_fabo}. We notice from the table that our model does a good job extracting the underlying emotion from the posture. From this table, we can see there is a mismatch between face and posture emotion recognition accuracies which means not all samples are getting similar output labels from both models. Further analysis of the results shows that there is around 12.8\% sample that gets different class label output from face and posture recognition. Using our re-weighting and retraining algorithm we could reduce the mismatch close to 5.8\%. Further reduction is hard to achieve and we attribute this fact to the idea that the face conveys more information than the posture.

\begin{table}[H]
\centering
\caption{Posture emotion recognition accuracy comparison on FABO dataset}
\begin{tabulary}{1.0\columnwidth}{|L|L|L|L|}
\hline
 Author& Model   &Face & Posture  \\ 
 \hline
  \cite{barros2015} &  CNN & 72.07 &57.8 \\
   \hline
  \cite{chen} & SVM with the RBF kernel&  66.5& 66.7 \\
  \hline
    \cite{gunes2009}& Adaboost    &35.2&73.2\\ 
\hline
   This work& Ensemble CNN    &92.7&83.1\\
\hline
\end{tabulary}
\label{p_fabo}
\end{table}

\subsection{Novelty Detection Cost}
We measured the average time it takes to perform classification using face and posture modalities. We tested them on the datasets mentioned earlier on an Intel 2.8 GHz Core I7 laptop computer without any GPU access. We found facial emotion recognition requires less time than posture based recognition. From table~\ref{noveltycost} we can see that adding posture analysis to detect novelty significantly increases processing time. So this is a trade-off that can be decided based on the systems need to handle novelty.  

\begin {table}[H]
\begin{tabular}{|c|c|}
\hline
\bf{Mode} & \bf{Time}\\
\hline
 Face &  1 \\ 
 \hline
 Posture& 1.5\\
 \hline
\end{tabular}
\caption{Average time taken by posture based model with respect to facial based model (values represents proportion).}
\label{noveltycost}
\end{table}

\section{Conclusion}

In this paper, we presented a NAERS-a novelty aware emotion recognition system that deals with novelty in a continuous learning fashion. We showed some of the early experiments with promising results. Some of the components on the system are yet to be completed and we are focusing on those as the next step of this work. 

\section{Acknowledgement}
This research is supported, in part, by the Defense Advanced Research Projects Agency (DARPA) and 3the Air Force Research Laboratory (AFRL) under the contract number W911NF2020003. The views and conclusions contained herein are those of the authors and should not be interpreted as necessarily representing the official policies or endorsements, either expressed or implied, of DARPA, AFRL, or the U.S. Government. We thank our team members on this project for all the discussions to develop this paper. Some of the ideas in this paper are based on our learning from the SAIL-ON meetings

\bibliography{main}
\end{document}